\documentclass[a4paper,conference]{IEEEtran}

\ifCLASSINFOpdf
\else
\fi
\usepackage{blindtext}
\usepackage{graphicx}
\usepackage{pifont}
\usepackage{multicol}
\usepackage{multirow}
\usepackage{hyperref}
\usepackage{makecell}
\usepackage{amssymb}
\newcommand{\cmark}{\ding{51}}%
\newcommand{\xmark}{\ding{55}}%
\usepackage{color} 
\newcommand{\etal}{\textit{et~al.}}
\usepackage{dblfloatfix}
\begin{document}
%
\title{Towards Automatic Parsing of Structured Visual Content through the Use of Synthetic Data}

\author{Lukas Sch\"olch\IEEEauthorrefmark{1}, Jonas Steinh\"auser\IEEEauthorrefmark{1}, Maximilian Beichter\IEEEauthorrefmark{1}, Constantin Seibold\IEEEauthorrefmark{2}, Kailun Yang\IEEEauthorrefmark{2},\\Merlin Knaeble\IEEEauthorrefmark{2}, Thorsten Schwarz\IEEEauthorrefmark{2}, Alexander Maedche\IEEEauthorrefmark{2}, and Rainer Stiefelhagen\IEEEauthorrefmark{2}
\\Karlsruhe Institute of Technology, Germany
\\\IEEEauthorrefmark{1}equal contribution, {\tt\small \{firstname.lastname\}@student.kit.edu}, \IEEEauthorrefmark{2}{\tt\small \{firstname.lastname\}@kit.edu}
}


%


\maketitle


\begin{abstract}
Structured Visual Content (SVC) such as graphs, flow charts, or the like are used by authors to illustrate various concepts.  While such depictions allow the average reader to better understand the contents, images containing SVCs are typically not machine-readable. This, in turn, not only hinders automated knowledge aggregation, but also the perception of displayed information for visually impaired people. 
In this work, we propose a synthetic dataset, containing SVCs in the form of images as well as ground truths. We show the usage of this dataset by an application that automatically extracts a graph representation from an SVC image. This is done by training a model via common supervised learning methods.
As there currently exist no large-scale public datasets for the detailed analysis of SVC, we propose the Synthetic SVC (SSVC) dataset comprising 12,000 images with respective bounding box annotations and detailed graph representations. Our dataset enables the development of strong models for the interpretation of SVCs while skipping the time-consuming dense data annotation. 

We evaluate our model on both synthetic and manually annotated data and show the transferability of synthetic to real via various metrics, given the presented application.
Here, we evaluate that this proof of concept is possible to some extend and lay down a solid baseline for this task. 
We discuss the limitations of our approach for further improvements.
Our utilized metrics can be used as a tool for future comparisons in this domain.
To enable further research on this task, the dataset is publicly available at \href{https://bit.ly/3jN1pJJ}{https://bit.ly/3jN1pJJ}.

\end{abstract}

\IEEEpeerreviewmaketitle

\section{Introduction}


\setlength{\abovecaptionskip}{0pt}
\setlength{\belowcaptionskip}{-30pt}

\begin{figure}[t]
    \centering
    \includegraphics[width=0.85\linewidth]{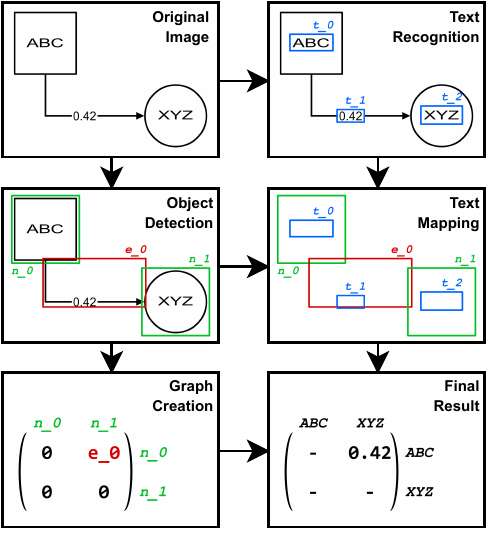}
    \caption{This flow chart describes the steps we used to extract a graph representation from an SVC image. At first, structures and texts are detected independently. Later they are merged again by using their bounding boxes and thus results in a complete representation of the SVC.}

    \label{fig:intro}
\end{figure}

An image says more than a thousand words. Following this line of thought, in almost all fields of living one tries to convey complex concepts in a visual manner. As such, fields like engineering or teaching cannot be imagined without the use of structured visual content (SVC) in the form of flow charts, diagrams, or similar graphs~\cite{mcgrath2005visual}. While SVC allow an immense condensation of information, it becomes hard to make these types of information machine parsable and therefore it is difficult to automatically aggregate information and enable various downstream tasks, such as accessibility, \textit{e.g.}, for visually impaired people. Thus, it is necessary to translate SVC into a machine parsable structure such as a graph representation to enable further processing.

The analysis of documents is a long established problem in computer vision. Most work, hereby, focuses either on the identification of text~\cite{smith2007overview,shi2015endtoend} or segmentation of different content types ~\cite{nguyen2021tablesegnet,SPaSe,WiSe}. As such Haurilet~\etal~\cite{SPaSe,WiSe} utilize deep learning methods to partition lecture slides into their specific parts to enable further downstream tasks. While such work addresses the identification and contextualization of SVC, the SVC itself remains untouched and cannot be interpreted. Auer~\etal~\cite{optical_graph_recognition} try to reconstruct the graph representation of a well defined subcategory of SVC through the use of classical computer vision algorithms which can be susceptible to error when applied to unconsidered domains. 

\setlength{\abovecaptionskip}{6pt}
\setlength{\belowcaptionskip}{-10pt}
\begin{figure*}[t]\label{gen_img}
\centering
\begin{tabular}{cc}
    \includegraphics[width=0.3\linewidth,height=0.23\linewidth]{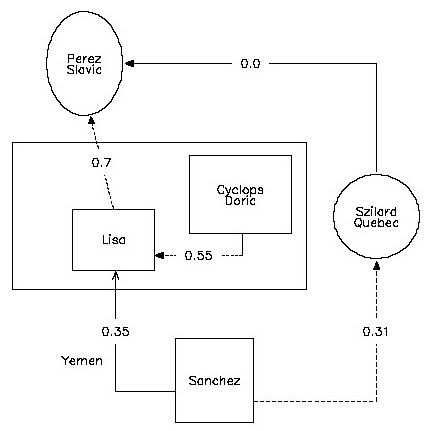}
    &
    \includegraphics[width=0.4\linewidth,height=0.23\linewidth]{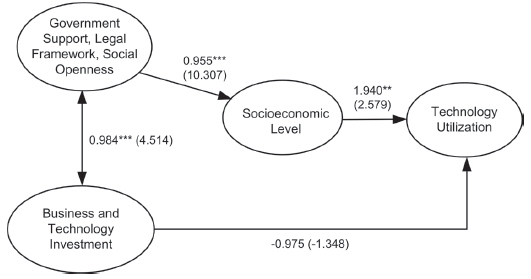}  \\
    \footnotesize{(a) Example of our generated dataset.} & 
    \footnotesize{(b) Example from the DISKNET dataset.}
\end{tabular}
\caption{Comparison of a generated image (a) with a real one from the DISKNET dataset (b).}
\label{fig:example_synthetic_vs_disknet}
\end{figure*}

We observe underlying similarities between various types of SVC. Nodes relating to objects are typically represented by geometrical structures such as  boxes, ovals, etc. in varying forms whereas their relationships are displayed via different types of arrows. Despite building upon simple structures, highly complex graphs can become too difficult to interpret for handcrafted methods. To combat this issue, we propose the use of a synthetic dataset in which we artificially generate random visual graphs to train deep learning methods for the detection of nodes and their relationships. Our approach can be seen in Fig.~\ref{fig:intro}. We generate a large scale SVC dataset and propose a method to automatically parse their graph representation. We evaluate our method not only on our synthetic dataset but also on the real world DISKNET~\cite{Disknet} dataset, which among others contains various SVC which were automatically extracted from different real world papers summarized in an extensive knowledge base, to highlight both its strengths and weaknesses.

\section{Related Work}

\subsection{Document Analysis}
\label{subsec:document_analysis}
Recent works in the field of document analysis seem to focus on one of two different problems. On the one hand, the area of text recognition aims to identify text in images and transform it to machine understandable text representations. This is for example done in \cite{smith2007overview} and \cite{shi2015endtoend} by methods of machine learning.  More advanced versions even try to recognize handwritten mathematical formulas \cite{maths_handwritten} or printed formulas \cite{math_printed}.
On the other hand, the focus is set on the recognition of document structures, often via segmentation. Examples are the page segmentation of lecture slide content \cite{WiSe} or the detection and segmentation of tables \cite{nguyen2021tablesegnet}.
In contrast to existing literature, we combine both text recognition and content localization. Thus, after the identification of nodes and edges we utilize text recognition models to deal with their content. We use this to build the graph representation of the SVC.

\subsection{Structured Visual Content}
\label{subsec:structured_visual_content}
In regards to our use case, only little work has been done on the recognition of SVC in general. Existent research focuses on similar topics, where the detection of structure in visuals is important. 
The recognition of mathematical formulas from text often uses a tree based structure to capture the structure and relation of different components within the formula \cite{math_graph}.
Some work focuses on graph structures and the associated construction of metrics. Le~\etal~\cite{LE2018118} focuses on the recognition of subgraphs in graph representations of comic book images with state of the art methods and also addresses the availability of ground truth datasets. Possibilities of calculating the edit distance of graphs efficiently are considered in~\cite{ABUAISHEH201796}.
More closely related to our definition of SVC are the following related works.
Building upon the real-world dataset (DISKNET), Scharfenberger~\etal~\cite{scharfenberger2021augmented} train a model for the extraction of such SVC. They are however limited by the low number of manually labeled training data available for their context. Further, their research still lacks in detecting node text content and in detection of edges and assigning the weights and their text. 
Awal~\etal~\cite{handwritten} try to extract a graph-like structure from hand drawn flow charts. This separates the task into two parts. Recognizing the elements that represent the graphical components and extracting the text itself.
Auer~\etal~\cite{optical_graph_recognition} on the other hand do not consider text and concentrate on recognizing limited graph structures. Those consist only of fully colored circular nodes and bidirectional straight edges. Their work uses a multi-step procedure with pre-processing, segmentation, topology recognition, and post-processing, while mainly using conventional computer vision methods and pixel based operations. 
Vasudevan~\etal~\cite{knowledge_extract} go one step further, their goal is to directly extract the knowledge from flow charts without recognizing an explicit graph structure. Similarly, Wu~\etal~\cite{code_extract} want to generate code from given flow charts. Both works make strict assumptions about the given images and focus strictly on their use case.

In contrast, our approach aims at SVC more in general and does not demand a certain type of graph structure. Therefore, we strive to detect basic elements such as nodes and their content, as well as different types of edges and their weights. The final graph representation should be purely descriptive, without being specialized for any certain use case. Further, we alleviate the issue of little available, ground-truth annotated data \cite{scharfenberger2021augmented} by combining manually labeled with synthetic data.

\section{Synthetic Structured Visual Content}

The creation of SVCs in this work is subject to various targets. We generate images of different sizes that present nodes and edges connecting them. Furthermore, groupings of nodes are possible and random text fields are included in these images. In addition, different design choices are applied to these objects, to increase the possible variations.

A node can be a rectangle or an ellipse and displays text. Multiple nodes can be framed by a rectangular grouping. Each node can be connected to one or multiple other nodes or to a grouping by an edge, but it does not have to. A grouping itself can only be connected to other nodes outside, but not to nodes within itself. An edge can be a straight line or an angled one, while being uni-directional or bi-directional. A variation of these edges is striped and the arrow tips can be filled or just line-work. Each edge has a numerical weight, presented by a text near the line, sometimes with an additional white background, which could overlap with the edge itself. All variations mentioned are distributed uniformly. Furthermore, different text sizes are used in these images, as well as different line widths and drawing strategies such as anti-aliasing.

In our implementation, we used a grid structure to place the objects. At a random position within each cell there may either be a node, a background text or nothing. Textual content is chosen from a list \cite{text_source}. Then a number of edges is placed between the nodes, but no connection is made twice, also avoiding collisions with nodes. Edge weights between 0 and 1 are annotated as well. Just as the variations in design, all positional decisions are afflicted by randomness. The size of the image, grid structure and nodes are all chosen randomly, as well as the number of nodes, edges and other elements. An example of such an SVC can be seen in Fig. \ref{fig:example_synthetic_vs_disknet}a.

\begin{figure}[]
    \centering
    \includegraphics[width=0.9\linewidth]{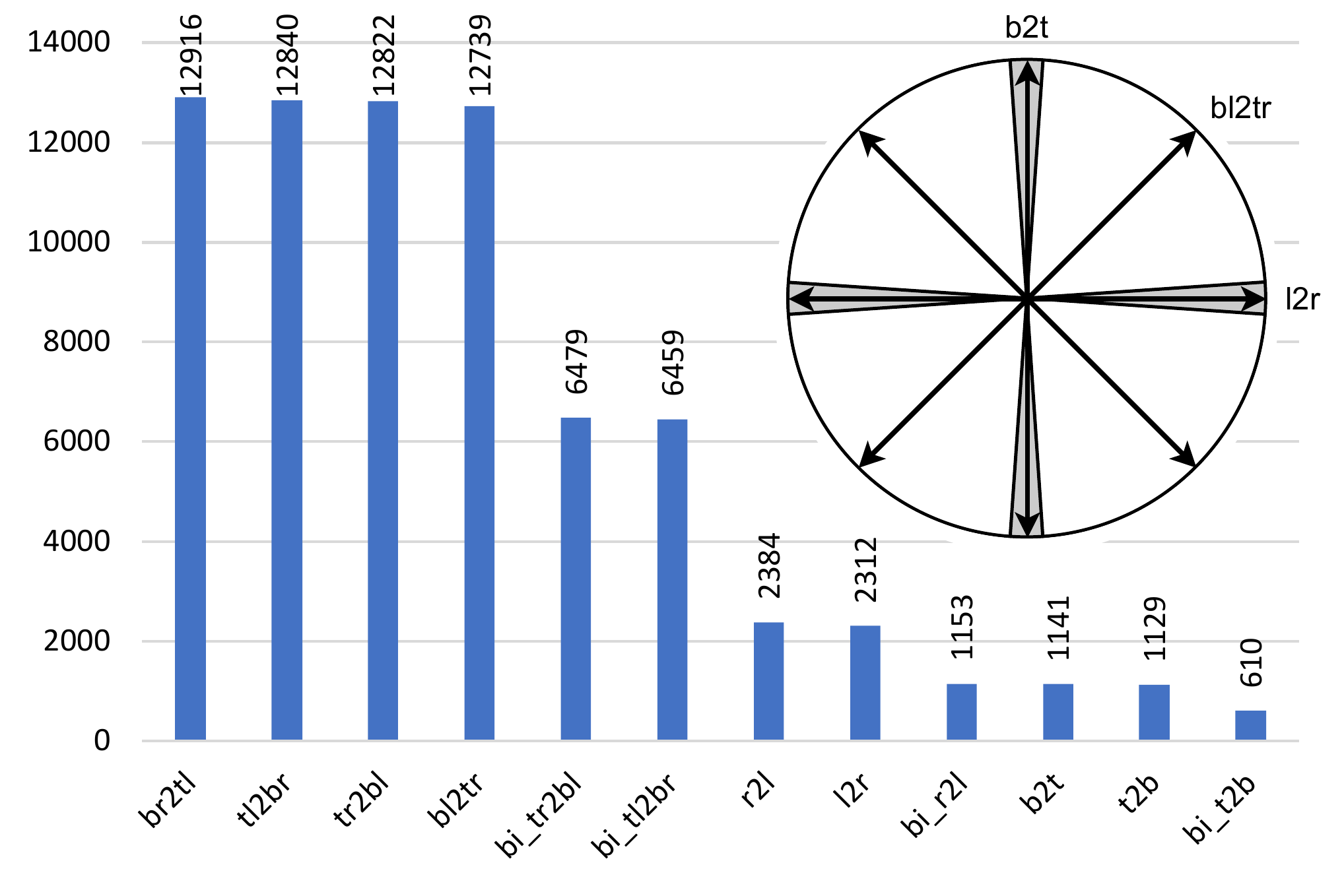}
    \caption{Statistics on Edge Classes}
    \label{fig:edges}
\end{figure}

Together with these images, we generate ground truth data for all tasks. 
For the object detection, bounding boxes are stored for the objects and separately for text segments.
Nodes of all variations and groupings are labeled ``node'', while edges are differentiated by their start and end point within a bounding box. In our case these can be in two opposing corners (diagonal edges) or at two opposing side centers (horizontal and vertical edges). 
This leads to 8 different edge types as seen in Fig. \ref{fig:edges}. 
To also represent bidirectional edges, which have two end points, we add further 4 classes in the same way.
Since we encountered problems with very small bounding boxes and under represented classes, we chose to label diagonal edges, that fall into +/-2.5 degrees of 0/90/180/270 degrees with the closest horizontal or vertical label.
We save bounding boxes for all texts within a node, besides an edge or in the background according to their purpose: ``node", ``edge", or ``plain".

For graph recognition tasks, we directly create a document that describes the generated graph structure. In this document, there is a list of all nodes, specifying their id, content and potential sub nodes. If a node has a sub node it is considered a grouping. Likewise, a list of all edges is defined with an id, weight, source-node and target-node id and information regarding the arrow tips for each edge. 


\setlength{\abovecaptionskip}{-6pt}
\setlength{\belowcaptionskip}{-30pt}

In total, we generated 12,000 images and ground truths, with a 70\%, 20\%, 10\% split into train, validation, and test data.

These images contain a total amount of 167,685 objects, 94,701 of them are nodes and 72,984 are edges. A detailed description of the edges and their respective classes is shown in Fig. \ref{fig:edges}. 
Diagonal edges are much more likely than horizontal or vertical edges. Furthermore, in our implementation, an edge has a 20\% chance to be bi-directional. This is visible in the statistics, as for example \#bl2tr + \#tr2bl $\approx$ 4 * \#bi\_tl2br. 
The naming of the edge classes represent their orientation within their bounding box. For example, ``tl2br" means ``top-left to bottom-right", likewise ``l2r" means ``left to right" referring to a horizontal edge pointing right (see Fig. \ref{fig:edges}).
Furthermore, a distinction is made between bi-directional and unidirectional edges. We display an exemplary synthetic image as well as a real-world image from the DISKNET dataset in Fig. \ref{fig:example_synthetic_vs_disknet}.

\section{Methodology}
Our goal is to extract the information from an image presenting SVC.
Therefore, we address the problem of mapping from an image showing a graph structure to a set of nodes and edges and their relation. In the following, we name this resulting set of nodes and edges graph representation. Specifically, our task describes the function
$$
Image \rightarrow \{Node\} \cup \{Edge\}.
$$

Nodes in our case are defined by their unique id and their textual content. Optionally, a node can contain a set of sub-nodes to define aggregations of nodes called groupings. Similar to nodes, edges are also defined by their id and their weights. In addition, an edge has a source and a target node.

Given the generated dataset, containing images of SVC, bounding box annotations of all objects and ground truth graph representations, the problem can be seen as supervised learning formulation. Instead of defining all possible variations by hand as needed in the traditional approach, machine learning techniques are applied to generalize over a large set of variations provided by the dataset.

The task is divided into different steps (see Fig. \ref{fig:intro}). 
The first receives an image and is supposed to detect all structural components such as nodes and edges. This is done via common object detection methods, utilizing the generated bounding box annotations.
In parallel, text recognition methods are applied, which extract the textual content from the image together with a position.
The next step uses the detected objects and reconstructs their structural relation by means of nearest neighbour approaches and heuristics.
Here, the goal is to create a graph structure describing the connection between nodes by the different edges and the grouping relations.
After that, the extracted textual information are mapped to their matching nodes and edges and added to the final graph representation.

\subsubsection{Object Detection}
We used a pretrained version of the YOLOv5 model \cite{yolov5} in its large configuration and fine tuned it on the training set of 8.400 images. We refrain from using flipping in the data augmentation step, as this would conflict with our definition of edge orientations. Furthermore, the Intersection over Union (IoU) threshold in training and detection should be set to zero, as overlapping bounding boxes are normal in this application. The result of this step are bounding boxes for each image, describing the position of nodes and different types of edges. However, we lose some information. By describing edges only via a bounding box and general direction, the exact course of an edge, is lost. While this may lead to problems when assigning the text to the edges it can be improved in the future.

\subsubsection{Text recognition}
Since text is an elementary component of graphs, it must also be included in the final graph representation. Because our object detection deals only with the detection of the structure, we used \textit{easy-OCR} \cite{EasyOCR} for the text. This text recognition works parallel to the object detection (see Fig. \ref{fig:intro}) and has no dependencies on it. It takes the same image as the object detection and returns a list of all detected texts, as well as their bounding boxes and positions. These are used later in the graph creation for the assignment.

\subsubsection{Create Graph Representation}
The results from object detection and text recognition are now merged into a graph representation. This can be divided into two main steps:
\noindent \paragraph{Step 1} Construction of the graph structure\\
In this process, groupings are detected at the beginning. To do this, the overlaps of all bounding boxes are determined. If a box is almost completely inside another, it is considered a subnode with the larger node as the grouping.
Afterwards, the nodes are connected via edges. For this, the \textit{connection points} are searched for an edge on its bounding box.
The connection points represent the start and end points of an edge, which are derived by the edge direction class. For a diagonal or angled edge these points are in the corner of a bounding box, for a horizontal or vertical edge, they are on the center of a side.
For each connection point the next adjacent node is searched and connected to the edge. 
In the end, all elements are connected accordingly and we have created a graph structure, that contains all previously detected objects.
\noindent  \paragraph{Step 2} Text mapping:
After the graph structure has been created, the texts have to be assigned to the corresponding elements in the graph.
First, the overlap to the node boxes is determined for each text bounding box. If the overlap is large enough, the text is considered as a part of the node and added to it.
The remaining texts must therefore belong to edges, or have been placed freely in the image.
To assign the edges, all possible edge bounding boxes are determined for each text bounding box, based on their overlap.
Then the most probable edge bounding box for the text is determined, based on all possible courses of an edge within its bounding box. All remaining texts are assumed to be background texts and will not be added to the graph representation.

\setlength{\abovecaptionskip}{-6pt}
\setlength{\belowcaptionskip}{-10pt}
\begin{figure}[t]
    \centering
    \includegraphics[width=0.90\linewidth]{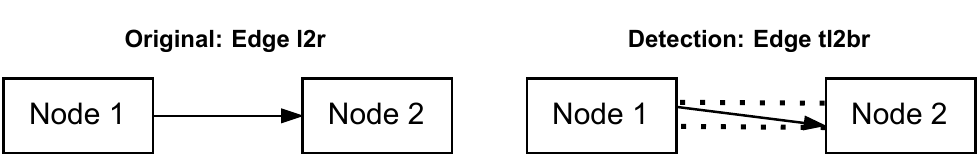}
    \caption{Slightly wrong edge classifications do not hinder Graph Reconstruction}
    \label{fig:edges_detection}
\end{figure}

\setlength{\abovecaptionskip}{-0pt}
\setlength{\belowcaptionskip}{-0pt}
\begin{table}[b]
\renewcommand{\arraystretch}{1.3}
\setlength{\tabcolsep}{4pt}
\caption{\#Labels and $mAP@[.5, .95]$ in Percent. \\ The edge names  follow the naming scheme of \{StartingPoint\}2\{EndPoint\}, \textit{i.e.} ``tl2br" for ``top left to bottom right". The prefix ``Bi" indicates bidirectional edges.}
\label{tab:map}
\centering
        \begin{tabular}{c|cc|cc|cc|cc|c}
        \multicolumn{10}{c}{\uppercase{\textbf{Synthetic Testset}}}
        \\
        \hline
        &\textbf{tl2br} & \textbf{br2tl}   & \textbf{tr2bl} & \textbf{bl2tr}   & \textbf{r2l}  & \textbf{l2r}   & \textbf{t2b}  & \textbf{b2t}  &  \textbf{nodes} \\
        \hline
        \hline
        \#Labels & 1248  & 1289                                         & 1294  & 1271                      & 254    & 227  & 114    & 110    &  9381 \\
        \hline
       $mAP$ &  86.7  & 87.3                                         & 87.8  & 86.3                      & 48.6   & 48.9 & 37.6   & 47.6   &  99.5 \\
        \hline 
        \hline
        & \multicolumn{2}{c|}{\textbf{bi\_tl2br}}    & \multicolumn{2}{c|}{\textbf{bi\_tr2bl}}   & \multicolumn{2}{c|}{\textbf{bi\_r2l}} & \multicolumn{2}{c|}{\textbf{bi\_t2b}}  &  \textbf{all} \\
        \hline
        \hline
        \#Labels & \multicolumn{2}{c|}{643}               & \multicolumn{2}{c|}{610}& \multicolumn{2}{c|}{120} & \multicolumn{2}{c|}{58}  &   16619 \\
        \hline
        $mAP$ & \multicolumn{2}{c|}{79.9}               & \multicolumn{2}{c|}{77.9}& \multicolumn{2}{c|}{47.0} & \multicolumn{2}{c|}{43.6}  &   67.6 \\
        \hline


        \multicolumn{10}{c}{\uppercase{\textbf{DISKNET Testset}}}
        \\
        \hline
        & \textbf{tl2br} & \textbf{br2tl}   & \textbf{tr2bl} & \textbf{bl2tr}   & \textbf{r2l}  & \textbf{l2r}   & \textbf{t2b}  & \textbf{b2t}  &  \textbf{nodes} \\
        \hline
        \hline
        \#Labels & 55    & 21                                           & 15    & 63                        & 2      & 39  & 7     & 7     &  210  \\
        \hline
        $mAP$ & 39.9  & 54.4                                         & 71.9  & 46.4                      & 15.0   & 0.8 & 0.0   & 0.0   &  87.5 \\
        \hline 
        \hline
        & \multicolumn{2}{c|}{\textbf{bi\_tl2br}}    & \multicolumn{2}{c|}{\textbf{bi\_tr2bl}}   & \multicolumn{2}{c|}{\textbf{bi\_r2l}} & \multicolumn{2}{c|}{\textbf{bi\_t2b}}  &  \textbf{all} \\
        \hline
        \hline
        \#Labels & \multicolumn{2}{c|}{1}                 & \multicolumn{2}{c|}{0}  & \multicolumn{2}{c|}{0}   & \multicolumn{2}{c|}{9}    &   429 \\
        \hline
        $mAP$ &  \multicolumn{2}{c|}{0.0}               & \multicolumn{2}{c|}{N/A}& \multicolumn{2}{c|}{N/A} & \multicolumn{2}{c|}{5.8}  &   29.2 \\
        \hline
        \end{tabular}
\end{table}

\section{Evaluation}

\label{sec:eval}
In this section, we evaluate the performance of object detection, generated graph structure and the text recognition. 

\subsection{Different Datasets}

To evaluate our pipeline we use four different datasets. We can divide these datasets into synthetic data and real data.

The synthetic dataset of 1,200 images is generated exactly as the training dataset and therefore has the same properties. 
From this dataset we take a subset of 30 clean, artefact-free images, as an adjusted version of the synthetic set.

For the real-world dataset, we manually annotated a subset from DISKNET~\cite{Disknet} containing 24 images. Again, we constructed an adjusted subset of 17 images, that do not include artifacts not available in our training set, such as curved edges.

\subsection{Metrics}

We rely on established metrics to evaluate our results from the object detection. Whereas, to evaluate the graph structure and the text recognition and assignments, we define our own metrics, since this is a problem without easy computable well established metrics. We overcome the time complexity of other metrics by exploiting the positional information.

\subsubsection{Metrics for the Object Detection}


To evaluate the object detection we use the $mAP$ defined as
$$
			    mAP = \frac{1}{|classes|}\sum_{c \in classes} \frac{\#TP(c)}{\#TP(c) + \#FP(c)},
$$


\noindent where a true positive ($TP$) is a prediction bounding box A that has an $IoU(A,B)$ to a groundtruth bounding box $B$,
$$
			    IoU(A, B) = \frac{A \cap B}{A \cup B},
$$
greater than a threshold, otherwise it is a false positive ($FP$). Furthermore, we evaluate the $mAP$ averaged for $IoU$ thresholds from $0.5$ to $0.95$ with a step size of $0.05$ (COCO’s \cite{coco} standard metric, simply denoted as $mAP@[.5, .95])$.

\subsubsection{Metrics for the Graph Structure}
Evaluating graphs and their isomorphism is a complex problem. After the prediction the result contains a graph representation with nodes and edges. But it contains no information about which structure of the predicted graph is related to which structure in the ground truth graph representation. This metric is calculated via the IoU of the predicted bounding boxes and the ground truth boxes of the synthetic or real world data. We define 
$$
    IsomorphicError =  \#mNodes + \#mEdges,
$$
with $mNodes$ and $mEdges$ as the amount of nodes and edges in the prediction and the ground truth representation without a match. Therefore, this number combines false negatives and false positives. Similarly we also define the normalized
$$
\quad NormIsomorphicError = \frac{IsomorphicError}{\# Nodes + \# Edges}
$$ 
to be able to better compare graphs of different size. In addition, we also calculated this error measure only for the nodes as well as the edges. We take only the missing nodes and the missing edges for the isomorphic error and normalize them with the number of nodes or edges in the graph. Note the $NormIsomorphicError$ can be larger than 1 as there could be more nodes or edges predicted than existing in the image.

\subsubsection{Metrics for the Text Mapping and the Text Recognition}

To compare the results of our text recognition as well as our assignment of the texts within the pipeline we define another error measure. We use the Levenshtein Distance \cite{Levenshtein}. The Levenshtein Distance $L(X,Y)$ is defined as the minimum number of deletions, insertions and replacement operations to change text $X$ to text $Y$. With this we can define our 
$$  
TextRecError = \frac{  \sum_{S_{matched}}L(GT_{content}, P_{content}) }{ \sum_{S_{matched}}\#GT_{content}},
$$ with $S_{matched}$ as every matched structure, $GT_{content}$ as the content of the structure in ground truth, $P_{content}$ as the content of the structure in prediction and the amount of characters in the ground truth content $\#GT_{content}$.

\subsection{Results}
In the following sections we present the results, divided into the object detection part, the graph representation and the text recognition as well as their assignment.

\begin{figure}[t]
    \centering
    \includegraphics[width=0.75\linewidth]{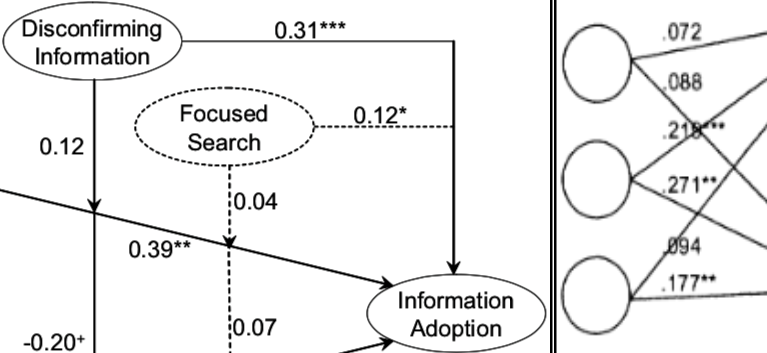}
    \caption{Example for human-created bad behavior (LEFT: merged edges; RIGHT: ambiguous weight positioning) graphs from the DISKNET dataset.}
    \label{fig:realErrors}
\end{figure}

\setlength{\abovecaptionskip}{-0pt}
\setlength{\belowcaptionskip}{-0pt}
\begin{table}[!b]
\renewcommand{\arraystretch}{1.3}
\caption{Normalized Isomorphic Error on the different Datasets. \\ Dir. $\widehat{=}$ if the graph is taken as a directed/undirected Graph.\\ BC. $\widehat{=}$   best Case $\quad $ WC. $\widehat{=}$  worst Case.}
\label{iso_error}
\centering
        \begin{tabular}{c||c||c|c|c|c|c}
        \hline
         & \bfseries Structure & \bfseries Dir. &  \bfseries Mean & \bfseries Std.& \bfseries BC. & \bfseries WC. \\ 
        \hline\hline
        \multirow{5}{*}{\rotatebox[origin=c]{90}{\bfseries \makecell{Synthetic \\ Testset}}} & \multirow{2}{*}{Full Graph}
         &\cmark  & 0.134  & 0.132 & 0 & 0.833\\
        & &\xmark  & 0.081  & 0.115 & 0 & 0.833\\
        \cline{2-7}
        & \multirow{1}{*}{Nodes}
        & & 0.016  &  0.0479 & 0 & 0.5\\
        \cline{2-7}
        & \multirow{2}{*}{Edges} & \cmark   & 0.287  & 0.281 & 0 & 3\\
             & & \xmark & 0.168  &  0.244 & 0 & 3\\
        \hline
        \hline
        \multirow{5}{*}{\rotatebox[origin=c]{90}{\bfseries \makecell{Adjusted \\ Synthetic \\ Testset}}} & \multirow{2}{*}{Full Graph}
         &
         \cmark  
         &  0.043 & 0.057 & 0.0 & 0.2 \\
        & &\xmark & 0.03 & 0.053 & 0.0 & 0.2 \\
        \cline{2-7}
        & \multirow{1}{*}{Nodes}
        & & 0.009 & 0.035 & 0.0 & 0.143\\
        \cline{2-7}
        & \multirow{2}{*}{Edges} & \cmark   & 0.1 & 0.163 &  0.0 & 0.667\\
             & & \xmark & 0.068 & 0.152 & 0.0 & 0.667\\
        \hline
        \hline
        \multirow{5}{*}{\rotatebox[origin=c]{90}{\bfseries \makecell{Disknet \\ Testset}}} & \multirow{2}{*}{Full Graph}
         &\cmark  &  0.328 & 0.139 & 0.091 & 0.654 \\
         & &\xmark  & 0.297 & 0.122 & 0.091 & 0.556 \\
        \cline{2-7}
        & \multirow{1}{*}{Nodes} 
        &  & 0.037 & 0.076 & 0.0 & 0.273\\
        \cline{2-7}
              & \multirow{2}{*}{Edges} 
            & \cmark    & 0.613 & 0.233 & 0.143 & 1.0\\
             & & \xmark & 0.553 & 0.211 & 0.143 & 1.0\\
        \hline
        \hline
        \multirow{5}{*}{\rotatebox[origin=c]{90}{\bfseries \makecell{Adjusted \\ Disknet \\ Testset}}} & \multirow{2}{*}{Full Graph}
         &\cmark  & 0.325  & 0.139 & 0.091 & 0.654\\
        & &\xmark  & 0.291  & 0.123 & 0.091 & 0.556\\
        \cline{2-7}
        & \multirow{1}{*}{Nodes}
        & & 0.021  &  0.062 & 0.0 & 0.25\\
        \cline{2-7}
        & \multirow{2}{*}{Edges} & \cmark   & 0.622  & 0.226 & 0.2 & 1.0\\
             & & \xmark & 0.558  &  0.213 & 0.2 & 1.0\\
        \hline
        \end{tabular}
\end{table}

\subsubsection{Evaluation of the Object Detection}

We evaluated the object detection on the synthetic test set and the DISKNET test set, visible in Table \ref{tab:map}.
The nodes and the most common edge types are detected very well by the model, while the less common types are recognized not as good. The comparison to the real world data set however shows, that those edge types are also less common there or not even present at all. 
As seen in the table, horizontal and vertical edges pose a much harder problem for the object detection than the diagonal edges. This can be due to their generally smaller bounding box.
After manual evaluation we noticed, that most errors do not come from missed detections but from wrong classifications. This means that an edge from bottom to top is detected as an edge from bottom left to top right. It can be argued that such an error is less bad than a complete miss, since the graph reconstruction can still work correctly in those situations (see Fig. \ref{fig:edges_detection}).


    

\subsubsection{Evaluation of Graph Representations}




We evaluated the reconstructed graph representations on all four test sets, visible in Table \ref{iso_error}. The normalized isomorphic error is calculated in two versions, where the first one includes wrong directions of the edges in the error, while the second does not and only measures the undirected graph.
The nodes are again the easiest part and are recognised almost perfectly on average even in the real world dataset.
Edges however are more difficult, while the detection of their direction is a big part of their error.
Furthermore, the mapping between edges of the ground truth and the detection results is very difficult.
In this implementation, we used heuristics to approximate a one to one mapping, which obviously can lead to problems.
This also means that one wrong assignment of an edge to a node can lead to two errors, as it counts towards a missing edge on the correct node and a false edge on that node.
Nevertheless, in multiple cases the graph reconstruction was perfect and able to create the wanted graph representation.



        


\subsubsection{Evaluation of Text Mapping and the Text-Recognition}

We evaluated the text recognition and text assignment on all four test sets, visible in Table \ref{ocr_error}. 
We observe, that the text recognition is on average better in the real world data set than in the synthetic data set.
Notice, that the error could be high, when a node is not detected, and therefore its text is falsely mapped. This influence is higher if there are only few nodes in the image.
The text recognition however is not the main task of our work and it is modular and easy exchangeable.

       
            
    

\section{Discussion}
\label{sec:discussion}

\begin{table}[t]
\renewcommand{\arraystretch}{1.3}
\caption{Text Recognition Error \\ BC. $\widehat{=}$  best Case $\quad $ WC. $\widehat{=}$  worst Case.}
\label{ocr_error}
\centering
        \begin{tabular}{c||c|c|c|c}
        \hline
        \bfseries Dataset & \bfseries Mean& \bfseries Std. &\bfseries BC. & \bfseries WC. \\
        \hline\hline
        Synthetic Testset & 0.423 & 0.429 & 0.0 & 13.0 \\
        Adjusted Synthetic Testset & 0.29 & 0.115 & 0.091 & 0.474 \\
        Disknet Testset & 0.254 & 0.214 & 0.039 & 0.692 \\
        Adjusted Disknet Testset & 0.2 & 0.187 & 0.039 & 0.692 \\

        \hline
        \end{tabular}
\end{table}

Comparing the results of our evaluation is very difficult because there are no comparable works on this task. Existing works address problems such as recognising structures and assigning it to different classes, as \cite{scharfenberger2021augmented}, because of the different task they are not suitable for comparison.
The evaluation shows that nodes themselves can be detected very confidently, while edges pose some problems. In general, horizontal or vertical edges are harder than diagonal edges, which can be due to their smaller bounding box size, the fact that they are less common in the dataset or due to the observation, that they are not always missed but incorrectly classified. It can be argued, that an edge from left to right which is detected as an edge from bottom left to top right is less bad than a complete miss.

Another problem with the detection of edges is their direction (c.f. Table \ref{iso_error}). Apparently the arrow heads are too small to be confidently detected.
In the generation, some problems producing ambiguous graphs are not excluded, since they also represent existing challenges. In Fig. \ref{fig:realErrors} two examples of such real world difficulties are shown. 
Furthermore, we show that a model is capable of generalizing over a large dataset, can extract the core structures of SVCs and apply them, also to maybe less complex SVCs. Therefore it is no problem, if the training data is sometimes a bit harder than the later real world use case. We show further examples in the supplementary material.




In our synthetic to real approach we can divide the limitations into two categories. One deals with the influence of the synthetic dataset. The second focuses on the applied method. 

\label{sec:limitations}
\subsubsection{Dataset}
The synthetic dataset provides first good approaches to generate different graphs. However, there are many variations that can be added to this dataset. For example, round edges and further edge types are possible, together with additional class labels representing their directions. Also, multiple nested groupings are missing and more shapes of nodes would be a useful extension. The horizontal as well as the vertical edges seem to be underrepresented in the dataset, which leads to a worse recognition results of these. 

Furthermore, the dataset does not contain any additional information, such as legends. These would be an interesting special case, since they often show elements of the graph without having a semantic meaning for the graph itself.
In general, our synthetic set is randomly generated. Therefore, the layouts and structures are also random in contrast to diagrams created by humans. This results in the entire diagram being distributed much more evenly in the image than it is often the case in reality. In reality, you often find converging structures that have many edges running up to each other. Such structures are underrepresented in the training set.
For further improvements, it would be interesting to extend the synthetic dataset to be able to represent even more real world graphs.

\subsubsection{Methods}
One of our main methodological limitation is the heuristic assignment of text to edges. This constraint originates in the problem, that we only work with bounding boxes of edges and lost the information about their exact path. Additionally, real world images do not follow any rule on how to assign weights to edges and therefore it is difficult to design a general heuristic. Similarly, our nodes are detected and processed only by their bounding boxes too. Even though this did not lead to any problems now, it maybe would be better to switch from an object detection via bounding boxes to segmentation masks, especially for better edge detection.

\section{Conclusion}
By storing the graph representation of SVC, the contained information within the image can be more easily accessed and utilized by other processes.
The automated parsing of such graph structures from SVC, however, is a challenging problem. To approach this, we have generated a synthetic automatically generated dataset which shares structures with its real counterpart. Through the automated dataset generation we gather annotations required for the training of deep learning models and enable tasks such as the parsing of graph structures. For the latter, our method matches detected nodes and edges with their respective captions based on spatial coordinates. 

We show promising results not only on the synthetic data but also on real world examples. Furthermore, we delineate potential limitations of our approach and, thereby, demonstrate the feasibility of our approach (\textit{i.e.} generating synthetic data, training on the aforementioned, applying the model on real world data) for SVC. Our approach allows for easily scalable training of performant models with only little manual annotation effort for, \textit{e.g.}, real world data. The \href{{https://bit.ly/3jN1pJJ}}{dataset} is publicly available. 
 We expect that our work will further enable easier automated analysis of SVC.

\bibliographystyle{IEEEtran}
\bibliography{own}

\end{document}